\documentclass[runningheads]{llncs}
\usepackage[T1]{fontenc}
\usepackage{graphicx}
\usepackage{amsmath}
\usepackage[hidelinks]{hyperref} 
\usepackage[inkscapearea=page]{svg} 
\svgsetup{inkscapeversion=1.1.2}
\usepackage{multirow} 
\svgpath{{figures/}}
\graphicspath{{figures/}}
\usepackage{booktabs} 
\usepackage{graphicx} 
\usepackage{subcaption} 

\def\*#1{\mathbf{#1}}

\begin{document}
\title{\texttt{PULPo}: Probabilistic Unsupervised Laplacian Pyramid Registration}

\author{Leonard Siegert\inst{1} 
\and
Paul Fischer\inst{1,2} 
\and 
Mattias P. Heinrich\inst{3} 
\and 
Christian F. Baumgartner \inst{1,2}}

\authorrunning{L. Siegert et al.}

\institute{Cluster of Excellence -- ML for Science, University of Tübingen, Germany\\ 
\and
Faculty of Health Sciences and Medicine, University of Lucerne, Switzerland\\
\and
Medical Informatics, University of Lübeck, Germany}

\maketitle          
\begin{abstract}
\setcounter{footnote}{0}
Deformable image registration is fundamental to many medical imaging applications. Registration is an inherently ambiguous task often admitting many viable solutions. While neural network-based registration techniques enable fast and accurate registration, the majority of existing approaches are not able to estimate uncertainty.
Here, we present \texttt{PULPo}, a method for probabilistic deformable registration capable of uncertainty quantification. \texttt{PULPo} probabilistically models the distribution of deformation fields on different hierarchical levels combining them using Laplacian pyramids. This allows our method to model global as well as local aspects of the deformation field. 
We evaluate our method on two widely used neuroimaging datasets and find that it achieves high registration performance as well as substantially better calibrated uncertainty quantification compared to the current state-of-the-art\footnote{The code is available at \url{https://github.com/leonardsiegert/PULPo}.}.

\end{abstract}

\section{Introduction}
\label{sec:introduction}

Deformable registration aims to establish a non-linear mapping between pairs of unaligned images. It is a key component in numerous clinical tasks, including the alignment of pre- and postoperative scans, as well as the monitoring of disease progression over time. Traditionally, registration methods addressed this problem through optimization, whereby an energy function is minimised for each image pair~\cite{rueckert1999nonrigid}. In order to improve registration performance, the displacement field was often optimized on several hierarchically organised resolution levels allowing to capture global as well as local effects~\cite{lester1999survey}. Substantial research was dedicated to diffeomorphic registration which ensures invertability and smoothness of the deformation fields~\cite{ashburner2007fast,beg2005computing}.   

While traditional methods are still widely used, they are very time consuming, often requiring many minutes or hours to align a 3D image pair. Recently, neural networks have emerged as a powerful alternative to traditional approaches. They allow for exceptionally fast deformation prediction with high accuracy~\cite{balakrishnan2019voxelmorph,meng2022non}. Expanding upon the concept of hierarchies in traditional registration methods, a number of works introduced neural network-based diffeomorphic registration approaches which progressively estimate and combine the deformation fields at different resolution levels~\cite{mok2020large,meng2022brain}. These approaches have been shown to lead to performance improvements over non-hierarchical methods. 

\begin{figure}[ht]
    \centering
    \includegraphics[width=\textwidth]{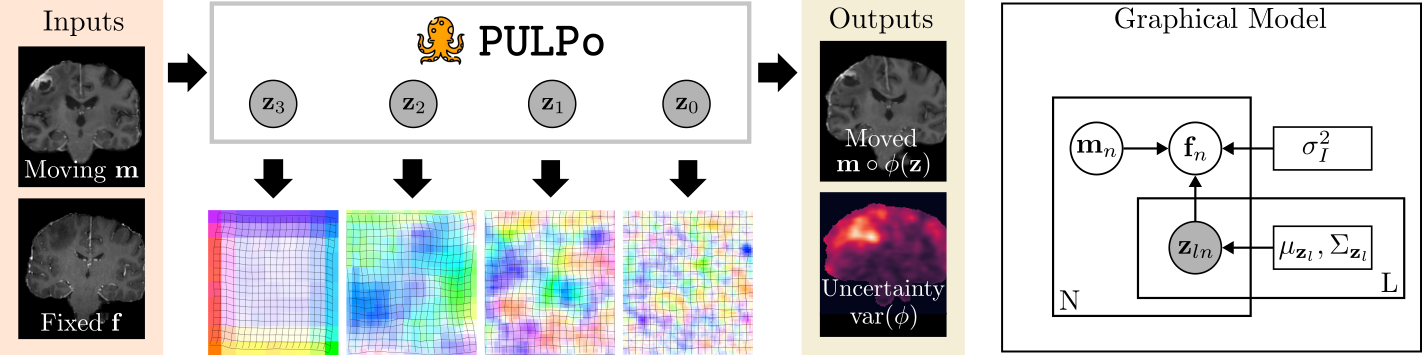}
    \caption{\textbf{Overview:} (Left) Our probabilistic unsupervised Laplacian pyramid registration approach (\texttt{PULPo}) hierarchically models the deformation field at different resolutions allowing for accurate registration and well-calibrated uncertainty quantification particularly in the presence of lesions. (Right) Graphical model for our method. }
    \label{fig:method-overview}
\end{figure}

An important but neglected area in registration is uncertainty quantification. The majority of existing approaches only output the most likely deformation. However, there may exist many viable alternative deformations that also lead to a good alignment~\cite{luo2019applicability}. The resulting aleatoric uncertainty in the deformation field is of high clinical significance for many applications such as disease monitoring or neurosurgery~\cite{risholm2010summarizing}. This problem is exacerbated when images contain unmatchable areas such as a preoperative scan containing a tumor and a follow-up scan where the tumor has been resected.

Early work on uncertainty quantification in traditional registration used probabilistic techniques to model the posterior of the deformation field given the input images, and used Markov-Chain Monte Carlo sampling to draw samples from this distribution~\cite{risholm2010summarizing,risholm2013bayesian,parisot2014concurrent,grzech2021uncertainty}. 

Uncertainty quantification in neural network-based approaches is currently understudied. Xu et al.~\cite{xu2022double} apply the popular Monte Carlo (MC) dropout approximation to get an estimate of epistemic uncertainty resulting from uncertainty in the model parameters. Recently, Chen et al.~\cite{chen2022transmorph} applied the same technique to a transformer-based registration architecture. Dalca et al.~\cite{dalca2018unsupervised,dalca2019unsupervised} proposed the probabilistic VoxelMorph which is an extention of their previously proposed approach~\cite{balakrishnan2019voxelmorph} by a variational inference of the deformation velocity field using a procedure inspired by conditional VAEs (cVAEs)~\cite{sohn2015learning}. Smolders et al.~\cite{smolders2022deformable} extended this approach for use in adaptive proton therapy. Since VoxelMorph uses a latent space matches the resolution of the input images, it may predispose the model to capture the distribution of local deformations at the expense of capturing the distribution of global deformations. Conversely, Krebs et al.~\cite{krebs2019learning} proposed a related method based on cVAEs, in which the latent space is in a low-resolution bottleneck of the architecture. While this approach allows to capture the distribution of local deformations, it may be restricted in capturing the distribution of local deformations. 
We note that Krebs et al. did not explore the uncertainty estimation aspect in their work.

Here, we propose a novel probabilistic unsupervised registration technique that can capture local as well as global deformations and their variability. The method uses a \textit{hierarchical} variational inference approach to probabilistically model the deformation field at increasingly fine resolution levels. For each level we estimate stationary velocity fields (SVFs)~\cite{ashburner2007fast}, which we then integrate to obtain diffeomorphic deformations. The deformation fields from the different resolution levels are then combined by leveraging the principle of Laplacian pyramids. 

Of the the neural network-based registration approaches capable of uncertainty estimation the probabilistic VoxelMorph~\cite{dalca2018unsupervised,dalca2019unsupervised} is arguably the most widely used due to its strong registration performance and the availability of easy-to-use, well-documented code. However, Grzech et al.~\cite{grzech2021uncertainty} observed that the probabilistic VoxelMorph produces uncertainty estimates with very small magnitudes, putting their utility into question. We confirm these findings in our own experiments in Sec.~\ref{sec:experiments}, and demonstrate that our probabilistic unsupervised Laplacian pyramid registration (\texttt{PULPo}) approach substantially outperforms the probabilistic VoxelMorph approach in terms of uncertainty quantification on the OASIS-1 and BraTS-reg datasets, while still producing well-aligned registrations. 

\section{Method}
\label{sec:method}

We model pairwise registration as a probabilistic generative process in which the fixed image $\mathbf{f}$ is generated by transforming the moving image $\mathbf{m}$ with the deformation field $\phi(\mathbf{z})$ that depends directly on a set of latent variables $\mathbf{z} = \{\mathbf{z}_l | l \in \{0, \dots, L-1\}\}$ as shown in Fig. \ref{fig:method-overview}. The latent variables $\mathbf{z}_l$ are arranged hierarchically, with the resolution of each level $l$ being half that of the level above ($l-1$). This allows them to capture local as well as global effects in the resulting deformation field. 

Following prior work~\cite{dalca2019unsupervised,krebs2019learning}, we define the likelihood of $\mathbf{f}$ as
\begin{equation}
    \label{eq:likelihood}
    p(\mathbf{f} | \mathbf{z}, \mathbf{m}) = \mathcal{N}(\mathbf{f}; \mathbf{m} \circ \phi(\mathbf{z}),\sigma_l^2 I) \ \text{,}
\end{equation}
where $\circ$ is the transformation of the moving image $\mathbf{m}$ with the deformation field $\phi$, and $\sigma_I^2$ denotes additive image noise. Our goal is to approximate the posterior distribution $p(\*z|\*f,\*m)$. As Maximum A Posteriori (MAP) estimation of the true posterior $p(\mathbf{z} | \mathbf{m}, \mathbf{f})$ is intractable, we use a variational approximation $q$ of the true posterior distribution, optimizing a lower bound to the true posterior \cite{kingma2013auto}. 

We define the prior $p(\mathbf{z})$ and variational posterior $q(\mathbf{z} | \mathbf{m}, \mathbf{f})$ of our generative model as products of Gaussians with diagonal covariance matrices  
\begin{equation}
    \begin{split}
    p(\mathbf{z}) &= \prod_{l=0}^{L-1} p(\mathbf{z}_l) = \prod_{l=0}^{L-1} \mathcal{N}(z_l; \mathbf{0}, I)\\
    q(\mathbf{z} | \mathbf{m},\mathbf{f}) &= \prod_{l=0}^{L-2} q(\mathbf{z}_l | \mathbf{z}_{l+1},\mathbf{m},\mathbf{f}) = \prod_{l=0}^{L-2} \mathcal{N}(\mathbf{z}_l; \mu_{\mathbf{z}_l|\mathbf{z}_{l+1},\mathbf{m},\mathbf{f}}, \Sigma_{\mathbf{z}_l|\mathbf{z}_{l+1},\mathbf{m},\mathbf{f}}) \ \text{,}
    \end{split}
\end{equation}
with $q(\mathbf{z}_{L-1} | \mathbf{m},\mathbf{f}) = \mathcal{N}(\mathbf{z}_{L-1}; \mu_{\mathbf{z}_{L-1}|\mathbf{m},\mathbf{f}}, \Sigma_{z_{L-1}|\mathbf{m},\mathbf{f}})$. 

To optimize the variational lower bound, we minimize the KL-divergence between the approximate posterior $q(\mathbf{z} | \mathbf{m},\mathbf{f})$ and the true
posterior $p(\mathbf{z} | \mathbf{m},\mathbf{f})$ leading to the following evidence lower bound (ELBO):
\begin{equation}
    \begin{split}
        \min \text{KL}[p(\mathbf{z} | \mathbf{m},\mathbf{f})||p(\mathbf{z} | \mathbf{m},\mathbf{f})] = \min \mathbf{E}_q[\log q(\mathbf{z} | \mathbf{m},\mathbf{f}) - \log p(\mathbf{z} | \mathbf{m},\mathbf{f})]\\
        = \min \sum_{l=0}^{L-1} \text{KL}[q(\mathbf{z}_l | \mathbf{z}_{l+1},\mathbf{m},\mathbf{f})|p(\mathbf{z}_l)] - \mathbf{E}_q[\log p(\mathbf{f}|\mathbf{z},\mathbf{m})] \ \text{.}
    \end{split}
\label{eq:main-objective}
\end{equation}
A detailed derivation can be found in Section A of the Supplemental Materials at the end of the document. 

We represent the deformation field using the SVF model~\cite{ashburner2007fast} where we first estimate a velocity field $\*v_l$ for each level, and then integrate it with scaling and squaring in 7 steps to yield an approximately diffeomorphic deformation field $\phi_l$.

\subsection{Neural Network Implementation}

\begin{figure}[t]
    \centering
    \includegraphics[width=\textwidth]{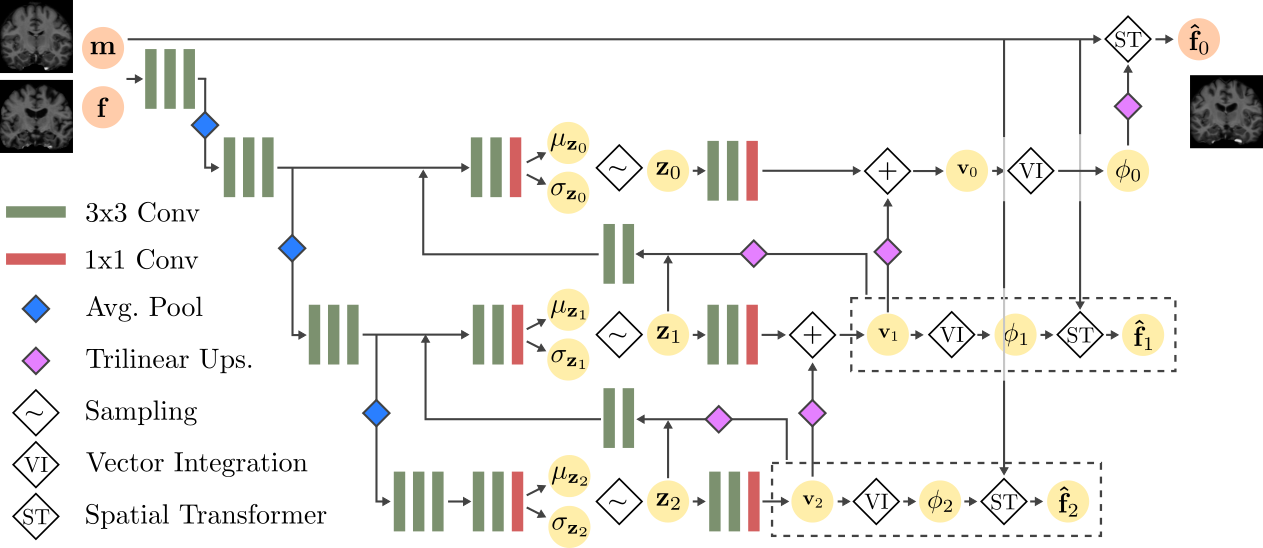}
    \caption{\textbf{Architecture:} Schematic network architecture with 3 latent levels. In practice, we used 4 latent levels. }
    \label{fig:architecture}
\end{figure}

We use neural networks to approximate $q(\*z| \*m, \*f)$ and the non-linear transformation $\phi(\*z)$. Following related work on hierarchical conditional VAEs~\cite{baumgartner2019phiseg,fischer2023uncertainty} the model is organized in $K$ total levels and $L<=K$ latent levels. A schematic of the architecture is shown in Fig.\,\ref{fig:architecture}. We used $K=5$ and $L=4$ for all experiments. The encoder predicts the parameters $\mu_{\mathbf{z}_l}$ and $\sigma_{\mathbf{z}_l}$ of each $\*z_l$ based on the moving ($\*m$) and fixed ($\*f$) input images. A sampling layer samples $\*z_l$ according to the predicted distribution. Those samples are then transformed into velocity fields $\*v_l$ for each level. The velocity fields processed by a vector integration layer to obtain deformation fields $\phi_l$ for each level, which are then transformed to the moved image $\*f_l$ using a spatial transformer layer. The outputs $\phi_0$ and $\*f_0$ are the final deformation field and moved image and have the same resolution as the input images. The outputs $\*f_l$ of the lower levels are used for deep supervision. Each level also informs the level above with a feedback connection containing a concatenation of $\*z_l$, $\*v_l$, $\phi_l$ and $\hat{\mathbf{f}}_l$.

Rather than independently estimating the velocity field $\*v_l$ for each resolution level, the velocity fields are incrementally added to the previously estimated velocity fields of the lower levels. This emulates the mechanics of a Laplacian pyramid \cite{burt1987laplacian}, and allows a gradual refinement of the final deformation akin to traditional hierarchical registration approaches. The vector integration on each level ensures that the resulting deformation fields are approximately diffeomorphic.

\subsection{Training}
\label{sec:training}

Since the loss terms in Eq.\,\ref{eq:main-objective} contain a sum over all voxels, higher levels are weighted higher in the resulting loss function since they contain more voxels. To account for this difference, we scale the losses on each level with $w_l = 2^{3l}$. 

We substitute the MSE loss resulting from the log-likelihood term in Eq.\,\ref{eq:likelihood} with the normalized cross-correlation (NCC)~\cite{avants2011reproducible}) with window size $k_l = 1+2(L-l)$ and a balancing factor $\gamma=0.05$. In preliminary experiments we observed a substantially better performance with the NCC-loss and used it for all experiments. We also use an auxiliary deep similarity loss to incentivize the method to use the lower levels. For this, we downscale the moving image $\mathbf{m}$ to the same resolution as the samples $\mathbf{z}_l$ and deformation fields $\phi(\mathbf{z}_l)$, and computed the NCC-loss with the level predictions $\hat{\mathbf{f}}_l = \mathbf{m}_l \circ \phi(\mathbf{z}_{l})$. Lastly, to ensure smooth deformation fields, we use a diffusion regularizer on the spatial gradients of the deformation field $\phi_l$ on each level. The total training objective for our model can be written as 
\begin{equation}
    \label{eq:training-objective}
    \begin{split}
    \mathcal{L} = \beta \sum_{l=0}^{L-1} w_l \text{KL}[q(\mathbf{z}_l | \mathbf{z}_{l+1},\mathbf{m},\mathbf{f})|p(\mathbf{z}_l)] \ -& \  \frac{\gamma}{\sigma_I^2} \sum_{l=0}^{L-1}w_l NCC(\mathbf{m}_l,\mathbf{f}_l, k_l) \\
    +& \ \lambda \sum_{l=0}^{L-1} w_l \sum_{v}^{V_l} ||\nabla \phi_v||^2 \ \text{.}
    \end{split}
\end{equation}
We used $\beta=0.1$ and $\lambda=0.025$ for our experiments. We set $\sigma_I^2 = 0.25$ on the highest level and $\sigma_I^2 = 1$ on all others.

\subsection{Inference}
\label{sec:inference}
After training, stochastic samples of the deformation field $\phi$ can be obtained with repeated forward passes through our model. This allows us to calculate the variation of the deformation field $\text{var}(\phi(\*z))$. Transforming the moving image $\*m$ with the samples further allows us to obtain an uncertainty estimate on a voxel level, i.e. $\text{var}(\*m \circ \phi(\*z))$. 

While the prediction could be obtained by averaging multiple samples, we propose an accelerated strategy: We perform a single forward pass, but instead of sampling $\mathbf{z}_l$, we directly propagate the predicted mean $\mu_{\mathbf{z}_l}$. Since the variational posterior $q$ is Gaussian when fixing the conditions, this procedure can be shown to approximate the MAP of the true posterior. 

\section{Experiments \& Results}
\label{sec:experiments}

We performed experiments on two 3D brain datasets and evaluated registration as well as uncertainty quantification performance. 

\subsection{Data}
\label{sec:data}

\subsubsection{OASIS-1}
We used the OASIS-1 3D brain MRI dataset \cite{hoopes2021hypermorph,marcus2007open}, containing both subjects with Alzheimer's disease and healthy controls, aged 18 to 96. We separated the subjects into a 354/30/30 train/validation/test split. The images were skull-stripped, bias-corrected and affinely registered. The resolution was isotropic 1mm resulting in volume sizes of 160 x 224 x 192. The dataset also contains segmentation maps with 35 anatomical labels, which we used to calculate the DSC metric. We used the anatomical landmarks from Taha et al. \cite{taha2023magnetic}. We used only one scan per subject and employed a pairwise inter-subject training scheme.

\subsubsection{BraTS-Reg}
We also used the BraTS-Reg 2022 dataset \cite{Baheti2021TheBT}, consisting of pre-operative and follow-up brain MRI scans of the same patient diagnosed with glioma. We used only the contrast-enhanced T1-weighted (T1CE) MR images, as Meng et al. \cite{meng2022brain} found T1CE-sequences to yield better performance than T1, T2, FLAIR, and multi-channel inputs. The data was split into training/validation/test following a 120/20/20 ratio. Landmarks in baseline and follow-up scans were manually annotated by clinical experts. After preprocessing, the volumes had an isotropic voxel size of 1mm and dimensions 144×192×160. We used a pairwise intra-subject training scheme, aligning the follow-up to the pre-operative scan.

\subsection{Metrics}
\label{sec:metrics}
We evaluated alignment in terms of the image intensity-based Root Mean Squared Error (RMSE) and the Target Registration Error (TRE) of the anatomical landmarks. For OASIS, we also evaluated the soft Dice Similarity Coefficient (DSC) of the anatomical segmentation maps.
We further evaluated the percentage of negative voxels of the Jacobian determinant (JD) of the deformation fields $\% |J_\phi| \leq 0$ as a measure of smoothness and invertibility. 

To evaluate the uncertainty estimation, we computed the NCC between the variance of the output and mean squared error
over $N=20$ samples~\cite{fischer2023uncertainty,baumgartner2019phiseg}. This provides a measure for the calibration, i.e. how well the method allocates uncertainty where it makes mistakes \cite{laves2021recalibration}. We report the NCC for the output image intensities ($NCC_{VX}$) and the anatomical landmarks ($NCC_{LM}$).

\subsection{Experiments}

We trained our proposed \texttt{PULPo} method as described in Sec.~\ref{sec:training}. 

We used the probabilistic diffeomorphic version of VoxelMorph (DIF-VM)~\cite{dalca2019unsupervised} as baseline method. Building on the authors' publicly available code, we implemented the probabilistic extension as described by Dalca et al. \cite{dalca2019unsupervised}. We used the same hyperparameters that Dalca et al. used in their experiments.

We trained DIF-VM using our NCC loss instead of the MSE as orginally proposed, since we found in preliminary results that the NCC version was superior across all metrics.

In order to better understand the effect of the probabilistic hierarchical structure of our method, we performed a non-hierarchical ablation of our method, \texttt{PULPo}$_{NH}$. While the base architecture was the same as that of the full \texttt{PULPo}, probabilistic sampling was only enabled on the highest level ($l=0$), and lower levels didn't contribute to the final deformation field. The losses on all other levels were set to zero, and $\mu_{\mathbf{z}_l}$ was propagated deterministically directly instead of sampling a $\*z_l$.

\subsection{Evaluation of Registration Performance}
\label{sec:performance_experiment}

\begin{table}[t!]
    \centering
    \caption{\textbf{Results:} Registration performance and uncertainty quantification metrics for all methods. }
    \resizebox{\textwidth}{!}{%
    \begin{tabular}{llcccc|cc}
    \textbf{Dataset} & \textbf{Method} & \textbf{RMSE} & \textbf{DSC} & \textbf{$\%|J_\phi|_{\leq 0}$} & \textbf{TRE} & \textbf{NCC$_{VX}$} & \textbf{NCC$_{LM}$}\\
    \hline
    \multirow{3}{*}{\textbf{OASIS-1}}& DIF-VM       & 0.039 & \textbf{0.804}  & \textbf{0.002}  & \textbf{2.943} & 0.210 $\pm$ 0.015 & -0.171 $\pm$ 0.201\\
    & \texttt{PULPo}$_{NH}$ & \textbf{0.036}  & 0.770  & 0.084   & 3.202 & 0.510 $\pm$ 0.051 & 0.092 $\pm$ 0.247\\
    & \texttt{PULPo}  & \textbf{0.036}  &  0.777  &  0.068 &  4.097 & \textbf{0.533} $\pm$ 0.039  & \textbf{0.302} $\pm$ 0.110\\
    \hline
    \multirow{3}{*}{\textbf{BraTS-Reg}}& DIF-VM       & \textbf{0.044} & - &  $\mathbf{<10^{-6}}$  & \textbf{2.321} & 0.264 $\pm$ 0.043 & -0.016 $\pm$ 0.366 \\
    & \texttt{PULPo}$_{NH}$ & 0.055  & - &  $\mathbf{<10^{-6}}$ & 3.030  & 0.474 $\pm$ 0.089 & \textbf{0.231} $\pm$ 0.352 \\
    & \texttt{PULPo}  & \textbf{0.044}    & - & 0.001 & 2.434 & \textbf{0.497} $\pm$ 0.073  & 0.229 $\pm$ 0.324 \\
    \end{tabular}}
    \label{tab:all-metrics}
\end{table}

\begin{figure}[ht]
    \centering
    \includegraphics[width=\textwidth]{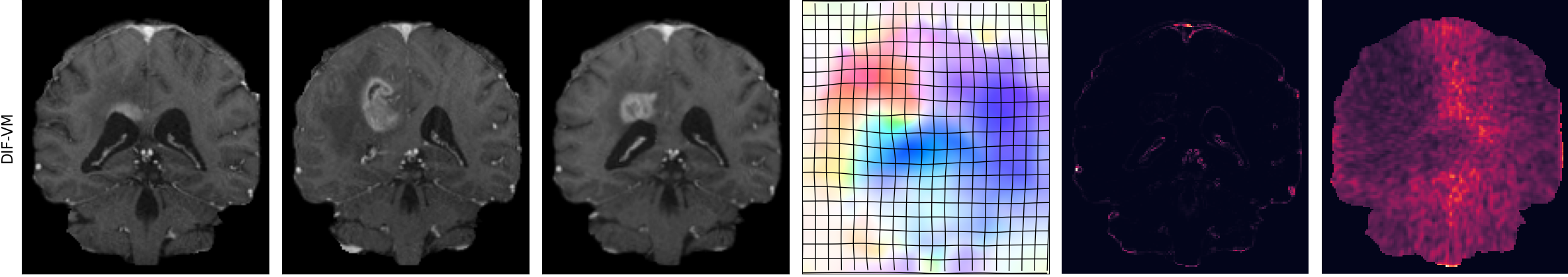}
    \includegraphics[width=\textwidth]{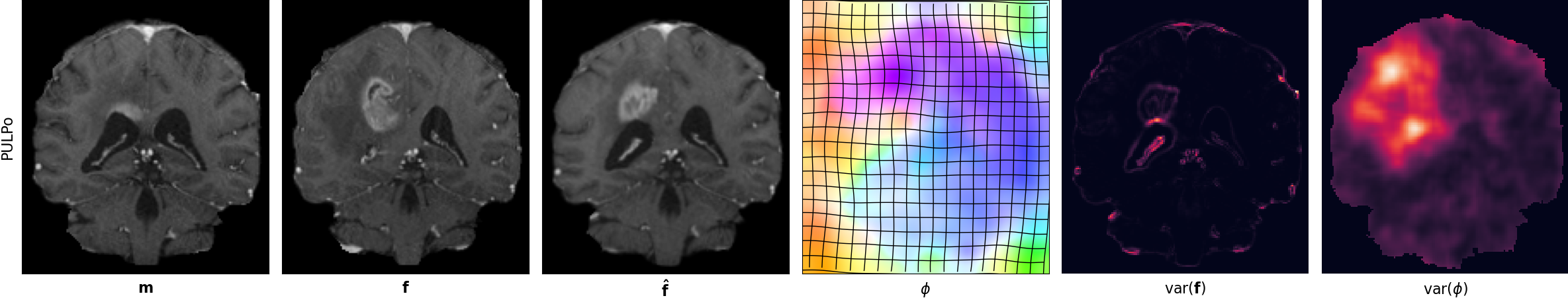}
    \caption{\textbf{Registration Results:}
    The figure shows the moving image $\mathbf{m}$, the fixed image $\mathbf{f}$, the prediction $\hat{\mathbf{f}}$, the predicted DF $\phi$, the voxel variance $\text{var}(\mathbf{f})$ and the DF variance $\text{var}(\phi)$ of the registration of T1CE-scan 5 from the BraTS-Reg dataset. The intensities of $\text{var}(\mathbf{f})$ and $\text{var}(\phi)$ needed to be visually amplified for DIF-VM due to their very small values. In this figure, the value ranges $\text{var}(\mathbf{f})$ for DIF-VM/PULPo are [0,3.98e-6]/[0,0.042], and the value ranges $\text{var}(\phi)$ for DIF-VM/PULPo are [0,3.37e-5]/[0,1.407].}
    \label{fig:experiments}
\end{figure}

We evaluated the registration performance in terms of the metrics described in Sec. \ref{sec:metrics}. The results in Tab.~\ref{tab:all-metrics} show that DIF-VM resulted in better alignment in terms of DSC and TRE, and smoother DFs in terms of \textbf{$\%|J_\phi|_{\leq 0}$}, however \texttt{PULPo} achieved better RMSE. Visual inspection of qualitative results (see Fig. \ref{fig:experiments}) suggested that both methods achieved good alignment with smooth deformations. Additional qualitative results are shown in Fig.~\ref{fig:supplemental_results} in Section B of the Supplementary Materials. 

The non-hierarchical ablation \texttt{PULPo}$_{NH}$ resulted in better alignment than \texttt{PULPo} in terms of the TRE, but in worse alignment in terms of the DSC on OASIS-1. On BraTS-Reg, \texttt{PULPo}$_{NH}$ resulted in significantly worse alignment than \texttt{PULPo}.

\subsection{Evaluation of Uncertainty Quantification}

The uncertainty calibration metrics NCC$_{VX}$ and NCC$_{LM}$ are shown in Tab.~\ref{tab:all-metrics} on the right. \texttt{PULPo} showed a substantially better calibration than DIF-VM, both in terms of image intensities (NCC$_{VX}$) and anatomical landmarks (NCC$_{LM}$) on both datasets. The negative NCC$_{LM}$ even indicates a negative correlation between landmark MSE and landmark uncertainty for DIF-VM. \texttt{PULPo} also generally performed better than the non-hierarchical ablation \texttt{PULPo}$_{NH}$ except on the BraTS-Reg landmarks, where the two methods performed similarly. This indicates that the probabilistic hierarchy allowed better calibrated uncertainty quantification.

We observed that DIF-VM produced samples with very little variance. This likely explains the poor performance on the uncertainty metrics. Upon closer inspection, we found that the $\sigma_\mathbf{z}$ of DIF-VM's random latent variable converged to a very small number making the model almost deterministic in behaviour. 
Visual inspection of qualitative results in Fig.~\ref{fig:experiments} confirmed that for \texttt{PULPo} the deformation field and image variance allocated large uncertainty around the tumor area where large uncertainties are expected. This can be further confirmed by inspecting the diversity of individual samples in Fig.~\ref{fig:supplemental_samples} in the Supplementary Materials. DIF-VM, on the other, hand produced a uniform DF variance map on which difficult to register areas cannot be discerned from others.

\section{Discussion and Conclusion}
\label{sec:conclusion}

Uncertainty quantification in medical image registration is of high importance for clinical applications such as disease monitoring or neurosurgery. However, the problem is severely understudied with only one prior work addressing this problem in the neural network setting. We have introduced a novel method for fast diffeomorphic 3D registration that can model global and local uncertainties by probabilistically modelling the deformation fields at different resolution levels, and combining them using a Laplace pyramid approach. We showed that our method produces accurate registrations as well as calibrated uncertainty quantification. 

Our experiments revealed a trade-off between accurate uncertainty estimation and high registration performance. While the probabilistic diffeomorphic voxelmorph (DIF-VM) achieved better registration performance on some metrics, its near deterministic behaviour severely limited its performance on uncertainty quantification. Future work will focus on further improving our architecture to improve registration performance, while maintaining well-calibrated uncertainty estimates.

\begin{credits}
\subsubsection{\ackname}  This work was supported by the Excellence Cluster 2064 ``Machine Learning --- New Perspectives for Science'', project number 390727645). The authors thank the International Max Planck Research School for Intelligent Systems (IMPRS-IS) for supporting Paul Fischer.

\subsubsection{\discintname}
The authors have no competing interests to declare. 
\end{credits}

\bibliographystyle{splncs04}
\bibliography{sources}

\begin{thebibliography}{10}
\providecommand{\url}[1]{\texttt{#1}}
\providecommand{\urlprefix}{URL }
\providecommand{\doi}[1]{https://doi.org/#1}

\bibitem{ashburner2007fast}
Ashburner, J.: A fast diffeomorphic image registration algorithm. Neuroimage  \textbf{38}(1),  95--113 (2007)

\bibitem{avants2011reproducible}
Avants, B.B., Tustison, N.J., Song, G., Cook, P.A., Klein, A., Gee, J.C.: A reproducible evaluation of ants similarity metric performance in brain image registration. Neuroimage  \textbf{54}(3),  2033--2044 (2011)

\bibitem{Baheti2021TheBT}
Baheti, B., Waldmannstetter, D., Chakrabarty, S., Akbari, H., Bilello, M., Wiestler, B., Schwarting, J., Calabrese, E., Rudie, J.D., Abidi, S.A.R., Mousa, M.S., Villanueva-Meyer, J.E., Marcus, D.S., Davatzikos, C., Sotiras, A., Menze, B.H., Bakas, S.: The brain tumor sequence registration challenge: Establishing correspondence between pre-operative and follow-up {MRI} scans of diffuse glioma patients. ArXiv  \textbf{abs/2112.06979} (2021), \url{https://api.semanticscholar.org/CorpusID:245131368}

\bibitem{balakrishnan2019voxelmorph}
Balakrishnan, G., Zhao, A., Sabuncu, M.R., Guttag, J., Dalca, A.V.: {VoxelMorph}: a learning framework for deformable medical image registration. IEEE transactions on medical imaging  \textbf{38}(8),  1788--1800 (2019)

\bibitem{baumgartner2019phiseg}
Baumgartner, C.F., Tezcan, K.C., Chaitanya, K., H{\"o}tker, A.M., Muehlematter, U.J., Schawkat, K., Becker, A.S., Donati, O., Konukoglu, E.: {PHiSeg}: Capturing uncertainty in medical image segmentation. In: Medical Image Computing and Computer Assisted Intervention--MICCAI 2019: 22nd International Conference, Shenzhen, China, October 13--17, 2019, Proceedings, Part II 22. pp. 119--127. Springer (2019)

\bibitem{beg2005computing}
Beg, M.F., Miller, M.I., Trouv{\'e}, A., Younes, L.: Computing large deformation metric mappings via geodesic flows of diffeomorphisms. International journal of computer vision  \textbf{61},  139--157 (2005)

\bibitem{burt1987laplacian}
Burt, P.J., Adelson, E.H.: The {Laplacian} pyramid as a compact image code. In: Readings in computer vision, pp. 671--679. Elsevier (1987)

\bibitem{chen2022transmorph}
Chen, J., Frey, E.C., He, Y., Segars, W.P., Li, Y., Du, Y.: {TransMorph}: Transformer for unsupervised medical image registration. Medical image analysis  \textbf{82},  102615 (2022)

\bibitem{dalca2018unsupervised}
Dalca, A.V., Balakrishnan, G., Guttag, J., Sabuncu, M.R.: Unsupervised learning for fast probabilistic diffeomorphic registration. In: Medical Image Computing and Computer Assisted Intervention--MICCAI 2018: 21st International Conference, Granada, Spain, September 16-20, 2018, Proceedings, Part I. pp. 729--738. Springer (2018)

\bibitem{dalca2019unsupervised}
Dalca, A.V., Balakrishnan, G., Guttag, J., Sabuncu, M.R.: Unsupervised learning of probabilistic diffeomorphic registration for images and surfaces. Medical image analysis  \textbf{57},  226--236 (2019)

\bibitem{fischer2023uncertainty}
Fischer, P., K{\"u}stner, T., Baumgartner, C.F.: Uncertainty estimation and propagation in accelerated {MRI} reconstruction. arXiv preprint arXiv:2308.02631  (2023)

\bibitem{grzech2021uncertainty}
Grzech, D., Azampour, M.F., Qiu, H., Glocker, B., Kainz, B., Loïc, L.F.: Uncertainty quantification in non-rigid image registration via stochastic gradient {Markov} chain {Monte Carlo}. In: MELBA Special Issue: Uncertainty for Safe Utilization of Machine Learning in Medical Imaging (UNSURE) 2020. vol. 12443, p.~3. MELBA (2021)

\bibitem{hoopes2021hypermorph}
Hoopes, A., Hoffmann, M., Fischl, B., Guttag, J., Dalca, A.V.: {HyperMorph}: Amortized hyperparameter learning for image registration. In: Information Processing in Medical Imaging: 27th International Conference, IPMI 2021, Virtual Event, June 28--June 30, 2021, Proceedings 27. pp. 3--17. Springer (2021)

\bibitem{kingma2013auto}
Kingma, D.P., Welling, M.: Auto-encoding variational bayes. arXiv preprint arXiv:1312.6114  (2013)

\bibitem{krebs2019learning}
Krebs, J., Delingette, H., Mailh{\'e}, B., Ayache, N., Mansi, T.: Learning a probabilistic model for diffeomorphic registration. IEEE transactions on medical imaging  \textbf{38}(9),  2165--2176 (2019)

\bibitem{laves2021recalibration}
Laves, M.H., Ihler, S., Fast, J.F., Kahrs, L.A., Ortmaier, T.: Recalibration of aleatoric and epistemic regression uncertainty in medical imaging. arXiv preprint arXiv:2104.12376  (2021)

\bibitem{lester1999survey}
Lester, H., Arridge, S.R.: A survey of hierarchical non-linear medical image registration. Pattern recognition  \textbf{32}(1),  129--149 (1999)

\bibitem{luo2019applicability}
Luo, J., Sedghi, A., Popuri, K., Cobzas, D., Zhang, M., Preiswerk, F., Toews, M., Golby, A., Sugiyama, M., Wells, W.M., et~al.: On the applicability of registration uncertainty. In: Medical Image Computing and Computer Assisted Intervention--MICCAI 2019: 22nd International Conference, Shenzhen, China, October 13--17, 2019, Proceedings, Part II 22. pp. 410--419. Springer (2019)

\bibitem{marcus2007open}
Marcus, D.S., Wang, T.H., Parker, J., Csernansky, J.G., Morris, J.C., Buckner, R.L.: Open access series of imaging studies (oasis): cross-sectional {MRI} data in young, middle aged, nondemented, and demented older adults. Journal of cognitive neuroscience  \textbf{19}(9),  1498--1507 (2007)

\bibitem{meng2022brain}
Meng, M., Bi, L., Feng, D., Kim, J.: Brain tumor sequence registration with non-iterative coarse-to-fine networks and dual deep supervision. arXiv preprint arXiv:2211.07876  (2022)

\bibitem{meng2022non}
Meng, M., Bi, L., Feng, D., Kim, J.: Non-iterative coarse-to-fine registration based on single-pass deep cumulative learning. In: International Conference on Medical Image Computing and Computer-Assisted Intervention. pp. 88--97. Springer (2022)

\bibitem{mok2020large}
Mok, T.C., Chung, A.C.: Large deformation diffeomorphic image registration with {Laplacian} pyramid networks. In: Medical Image Computing and Computer Assisted Intervention--MICCAI 2020: 23rd International Conference, Lima, Peru, October 4--8, 2020, Proceedings, Part III 23. pp. 211--221. Springer (2020)

\bibitem{parisot2014concurrent}
Parisot, S., Wells~III, W., Chemouny, S., Duffau, H., Paragios, N.: Concurrent tumor segmentation and registration with uncertainty-based sparse non-uniform graphs. Medical image analysis  \textbf{18}(4),  647--659 (2014)

\bibitem{risholm2013bayesian}
Risholm, P., Janoos, F., Norton, I., Golby, A.J., Wells~III, W.M.: Bayesian characterization of uncertainty in intra-subject non-rigid registration. Medical image analysis  \textbf{17}(5),  538--555 (2013)

\bibitem{risholm2010summarizing}
Risholm, P., Pieper, S., Samset, E., Wells~III, W.M.: Summarizing and visualizing uncertainty in non-rigid registration. In: International Conference on Medical Image Computing and Computer-Assisted Intervention. pp. 554--561. Springer (2010)

\bibitem{rueckert1999nonrigid}
Rueckert, D., Sonoda, L.I., Hayes, C., Hill, D.L., Leach, M.O., Hawkes, D.J.: Nonrigid registration using free-form deformations: application to breast {MR} images. IEEE transactions on medical imaging  \textbf{18}(8),  712--721 (1999)

\bibitem{smolders2022deformable}
Smolders, A., Lomax, T., Weber, D.C., Albertini, F.: Deformable image registration uncertainty quantification using deep learning for dose accumulation in adaptive proton therapy. In: International Workshop on Biomedical Image Registration. pp. 57--66. Springer (2022)

\bibitem{sohn2015learning}
Sohn, K., Lee, H., Yan, X.: Learning structured output representation using deep conditional generative models. Advances in neural information processing systems  \textbf{28} (2015)

\bibitem{taha2023magnetic}
Taha, A., Gilmore, G., Abbass, M., Kai, J., Kuehn, T., Demarco, J., Gupta, G., Zajner, C., Cao, D., Chevalier, R., et~al.: Magnetic resonance imaging datasets with anatomical fiducials for quality control and registration. Scientific Data  \textbf{10}(1), ~449 (2023)

\bibitem{xu2022double}
Xu, Z., Luo, J., Lu, D., Yan, J., Frisken, S., Jagadeesan, J., Wells~III, W.M., Li, X., Zheng, Y., Tong, R.K.y.: Double-uncertainty guided spatial and temporal consistency regularization weighting for learning-based abdominal registration. In: International Conference on Medical Image Computing and Computer-Assisted Intervention. pp. 14--24. Springer (2022)

\end{thebibliography}

\clearpage
\appendix
\section*{Supplementary Materials}

\renewcommand{\thesubsection}{\Alph{subsection}}

\subsection{Derivation of the Evidence Lower Bound (ELBO)}
In the following, we show how to derive the evide lower bound (ELBO) from Eq. 3 of the main text, starting from the KL-divergence between approximate posterior $q(\mathbf{z} | \mathbf{m},\mathbf{f})$ and the true posterior $p(\mathbf{z} | \mathbf{m},\mathbf{f})$:
\begin{equation}
    \begin{split}
        &\mathbf{KL}[q(\mathbf{z} | \mathbf{m},\mathbf{f})||p(\mathbf{z} | \mathbf{m},\mathbf{f}
        = \mathbf{E}_q[\log q(\mathbf{z} | \mathbf{m},\mathbf{f}) - \log p(\mathbf{z} | \mathbf{m},\mathbf{f})] \\
        = &\mathbf{E}_q\left[\log \prod_{l=0}^{L-1}q(\mathbf{z}_l | \mathbf{z}_{l+1},\mathbf{m},\mathbf{f}) - \log \prod_{l=0}^{L-1}p(\mathbf{z}_l | \mathbf{m},\mathbf{f})\right] \\
        = &\mathbf{E}_q\left[\log \prod_{l=0}^{L-1}q(\mathbf{z}_l | \mathbf{z}_{l+1},\mathbf{m},\mathbf{f}) - (\log \prod_{l=0}^{L-1}p(\mathbf{z}_l,\mathbf{m},\mathbf{f}) - \log p(\mathbf{m},\mathbf{f}))\right] \\
        = &\mathbf{E}_q\left[\log \prod_{l=0}^{L-1}q(\mathbf{z}_l | \mathbf{z}_{l+1},\mathbf{m},\mathbf{f}) - \log \prod_{l=0}^{L-1}p(\mathbf{z}_l,\mathbf{m},\mathbf{f})\right] + \log p(\mathbf{m},\mathbf{f}) \\
        = &\mathbf{E}_q\left[\log
        \prod_{l=0}^{L-1}q(\mathbf{z}_l | \mathbf{z}_{l+1},\mathbf{m},\mathbf{f}) - (\log p(\mathbf{f} | \mathbf{z},\mathbf{m}) + \log \prod_{l=0}^{L-1}p(\mathbf{z}_l) + \log p(\mathbf{m}))\right] + \log p(\mathbf{m},\mathbf{f}) \\
        = &\mathbf{E}_q\left[\sum_{l=0}^{L-1}[\log q(\mathbf{z}_l | \mathbf{z}_{l+1},\mathbf{m},\mathbf{f}) - \log p(\mathbf{z}_l)] - \log p(\mathbf{f} | \mathbf{z},\mathbf{m}) - \log p(\mathbf{m})\right] + \log p(\mathbf{m},\mathbf{f}) \\
        = &\sum_{l=0}^{L-1} \left[\mathbf{KL}(q(\mathbf{z}_l | \mathbf{z}_{>l},\mathbf{m},\mathbf{f})|p(\mathbf{z}_l))\right] - \mathbf{E}_q[\log p(\mathbf{f}|\mathbf{z},\mathbf{m})] + \log p(\mathbf{m},\mathbf{f}) - \log p(\mathbf{m}) \ \text{.}
    \end{split}
\end{equation}
As $\log p(\mathbf{m},\mathbf{f})$ and $\log p(\mathbf{m})$ are constant with respect to $\mathbf{z}$, they can not be optimized. The optimization
task is thus minimizing the Evidence Lower Bound (ELBO), consisting of the KL-divergence between the approximate posterior $q(\mathbf{z}_l | \mathbf{m},\mathbf{f})$
and the prior $p(\mathbf{z}_l)$, for each level $l$ and minimizing the
negative expectation of the log-likelihood $\log p(\mathbf{f}_l | \mathbf{m},\mathbf{z}_l)$:
\begin{equation}
\min \sum_{l=0}^{L-1} \text{KL}[q(\mathbf{z}_l | \mathbf{z}_{l+1},\mathbf{m},\mathbf{f})|p(\mathbf{z}_l)] - \mathbf{E}_q[\log p(\mathbf{f}|\mathbf{z},\mathbf{m})] \ \text{.}
\end{equation}

\clearpage
\subsection{Additional Figures}

\begin{figure}[ht]
    \centering
    \begin{subfigure}[b]{0.95\textwidth}
        \centering
        \includegraphics[width=\textwidth]{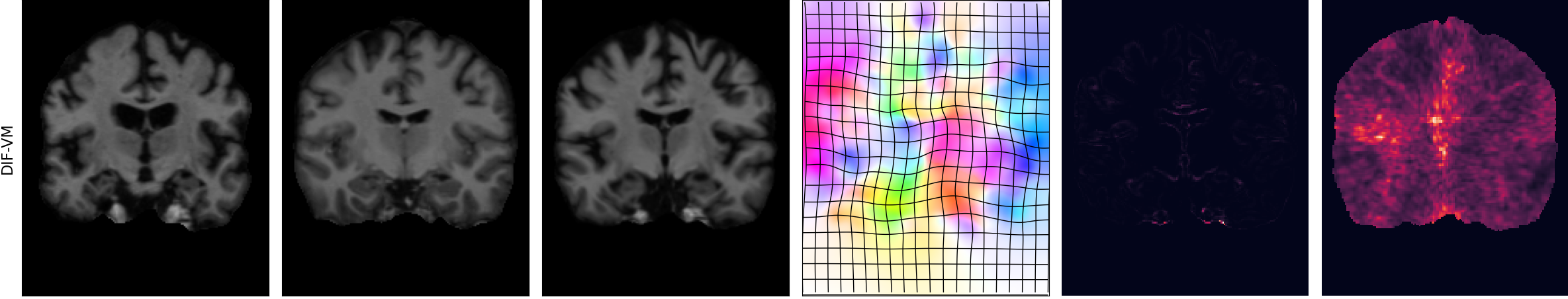}
        \includegraphics[width=\textwidth]{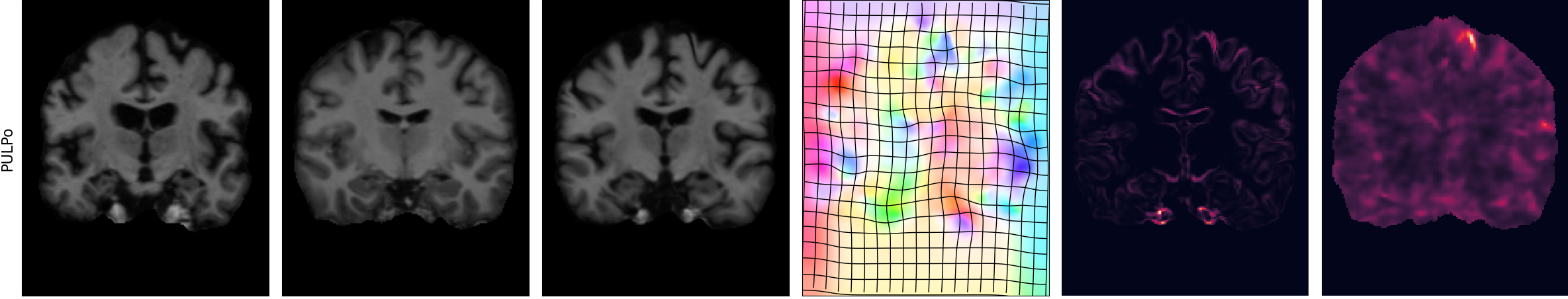}
        \caption{Example from OASIS}
        \label{fig:supplemental_results_oasis}
    \end{subfigure}

    \begin{subfigure}[b]{0.95\textwidth}
        \centering
        \includegraphics[width=\textwidth]{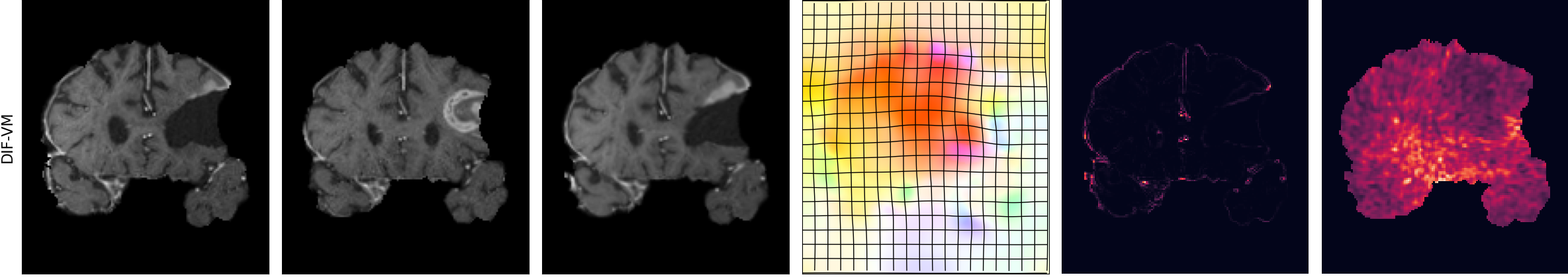}
        \includegraphics[width=\textwidth]{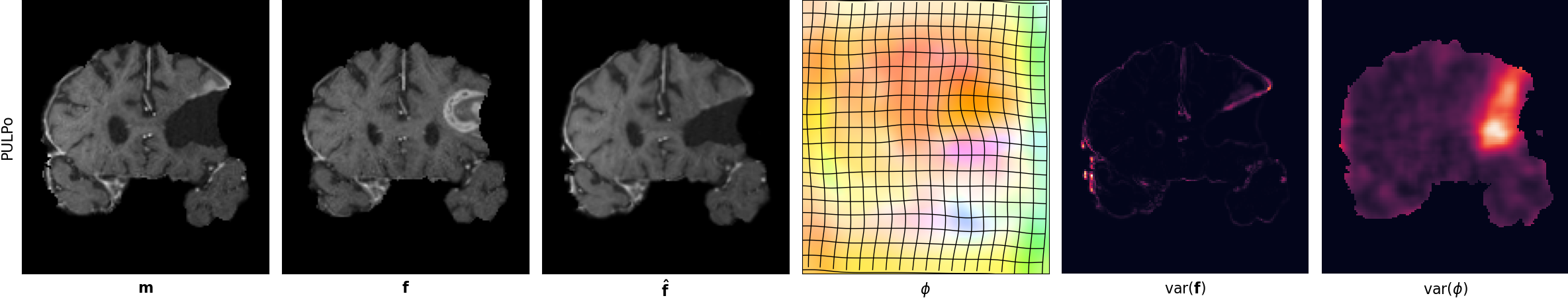}
        \caption{Example from BraTS-Reg}
        \label{fig:supplemental_results_brats}
    \end{subfigure}

    \caption{\textbf{Additional registration results:}
    The figure shows the moving image $\mathbf{m}$, fixed image $\mathbf{f}$, prediction $\hat{\mathbf{f}}$, predicted DF $\phi$, voxel variance $var(\mathbf{f})$ and DF variance $var(\phi)$ for two example registrations. (a) A coronal slice of the registration between subject 10 ($\mathbf{m}$) and subject 86 ($\mathbf{f}$) from OASIS. (b) A coronal slice of the registration for T1CE-scan 7.}
    \label{fig:supplemental_results}
\end{figure}

\begin{figure}[ht]
    \centering
    \begin{subfigure}[b]{0.95\textwidth}
        \centering
        \includegraphics[width=\textwidth]{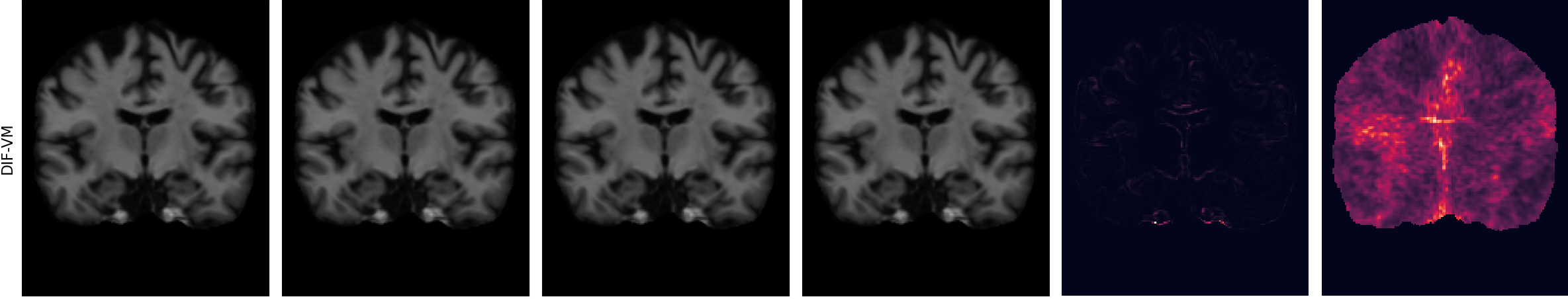}
        \includegraphics[width=\textwidth]{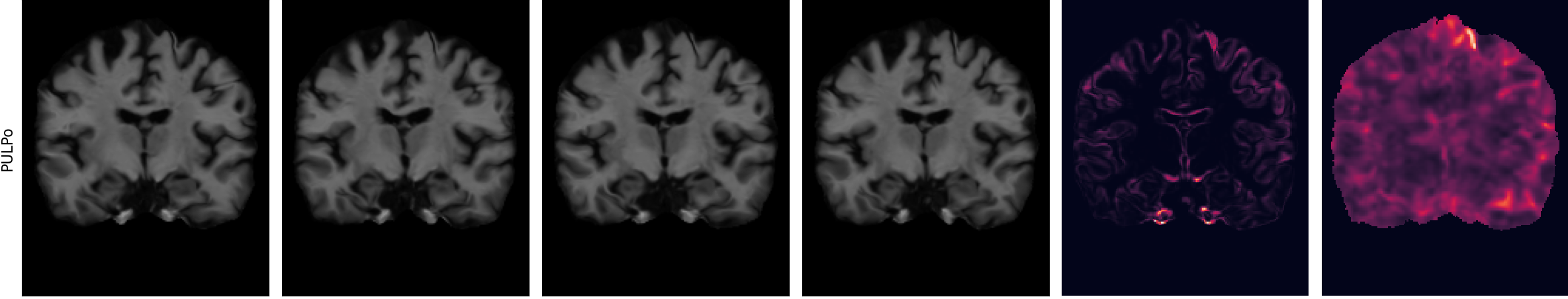}
        \caption{Samples for OASIS}
        \label{fig:supplemental_samples-oasis}
    \end{subfigure}
    \hfill
    \begin{subfigure}[b]{0.95\textwidth}
        \centering
        \includegraphics[width=\textwidth]{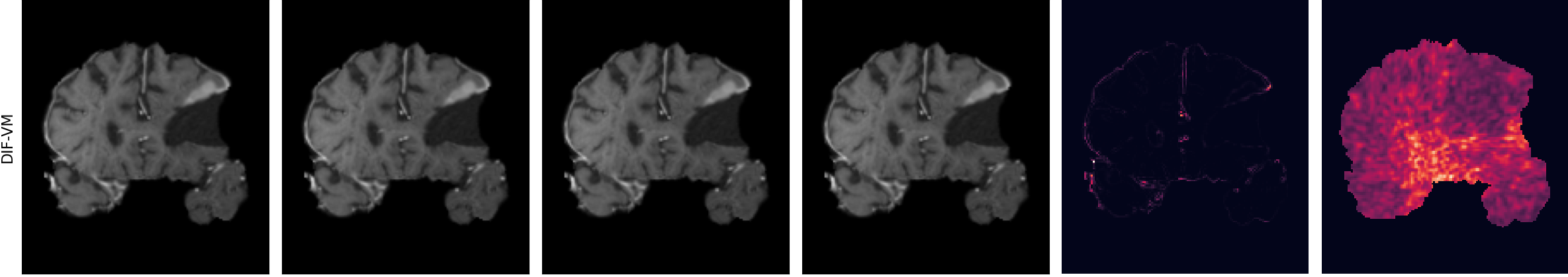}
        \includegraphics[width=\textwidth]{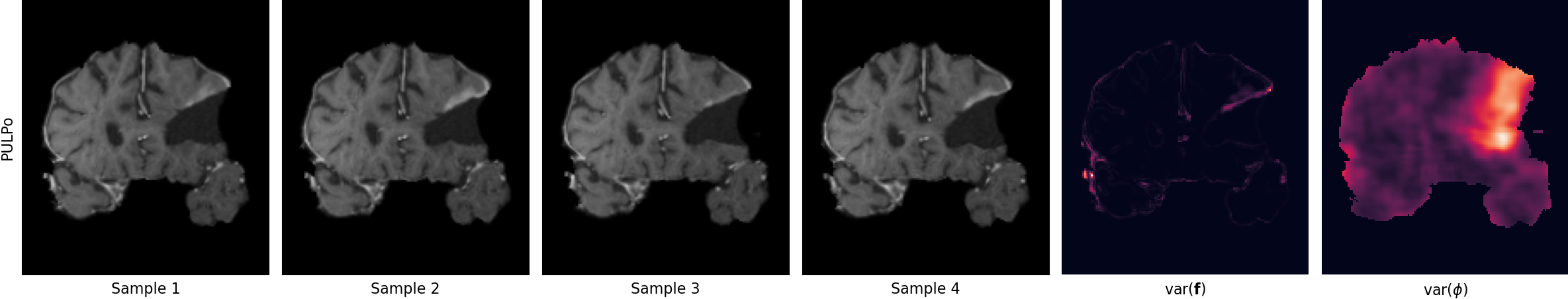}
        \caption{Samples for BraTS-Reg}
        \label{fig:supplemental_samples-brats}
    \end{subfigure}
    \caption{\textbf{Registration samples:} The figure shows four samples, the image variance, and the DF variance for an example registration. (a) A coronal slice of subject 10 ($\mathbf{m}$) and subject 86 ($\mathbf{f}$) on OASIS. (b) A coronal slice of T1CE-scan 7 on BraTS-Reg.}
    \label{fig:supplemental_samples}
\end{figure}

\end{document}